\documentclass[10pt,twocolumn,letterpaper]{article}

\usepackage{wacv}              % To produce the CAMERA-READY version
\definecolor{wacvblue}{rgb}{0.21,0.49,0.74}
\usepackage[pagebackref,breaklinks,colorlinks,allcolors=wacvblue]{hyperref}

\usepackage{multirow,color,colortbl,soul,bm,float,booktabs,amssymb,amsmath,amsfonts,wrapfig,xspace,caption}
\usepackage{algorithm}
\usepackage{algpseudocode}
\usepackage{tabularx}
\usepackage[misc]{ifsym}

\usepackage{etoolbox}
\newtoggle{withSupp}
\toggletrue{withSupp}

\usepackage[accsupp]{axessibility}
\usepackage{hyperref}
\usepackage{orcidlink}

\newcommand{\ours}{CrossPressor\xspace}
\newcommand\rurl[1]{%
\href{https://#1}{\nolinkurl{#1}}%
}

\definecolor{softred}{rgb}{1.0,0.6,0.6}
\definecolor{softorange}{rgb}{1.0,0.8,0.6}
\definecolor{softyellow}{rgb}{1.0,1.0,0.6}
\def\wacvPaperID{616} % *** Enter the WACV Paper ID here
\def\confName{WACV}
\def\confYear{2027}

\title{CrossPressor: Scalable Cross-View Fusion for Feed-Forward 3D Gaussian Splatting}

\author{
Yuxi Hu$^{1}$ \quad
Wenhao Dong$^{2}$ \quad
Jun Zhang$^{1}$ \quad
Friedrich Fraundorfer$^{1}$ \\
$^{1}$Graz University of Technology \qquad $^{2}$Ghent University \\
{\ttfamily\small
\{yuxi.hu, jun.zhang, friedrich.fraundorfer\}@tugraz.at
\{Wenhao.Dong\}@UGent.be
}
}
\begin{document}
\maketitle

%%%%%%%%% ABSTRACT
\begin{abstract}
Feed-forward 3D Gaussian Splatting (3D-GS) becomes increasingly difficult to scale to long multi-view sequences. Anchor-based compression methods such as ZPressor alleviate representation growth by decoding Gaussians only from a compact set of anchor feature maps, but keeping these anchors fixed forces each one to aggregate an increasing number of support views. We identify this representation--fusion mismatch and introduce \textbf{CrossPressor}, a training-free method for scalable anchor-based fusion. CrossPressor preserves all decoder-facing anchor queries while constraining the view-growing support sequence through coverage-cycled evidence routing, distributing retained spatial evidence across support views and fusion layers while keeping all support views represented. Routing before K/V projection, together with joint context packing, further reduces support-side fusion cost without retraining. CrossPressor maintains reconstruction quality close to dense fusion while achieving up to $26.12\times$ faster cross-view fusion and $1.43\times$ end-to-end acceleration on long input sequences.  Our project page is at: \url{https://yuhsihu.github.io/CrossPressor}.
\end{abstract}
%%%%%%%%% 

%%%%%%%%% BODY TEXT
\section{Introduction}
\label{sec:intro}

Feed-forward 3D Gaussian Splatting (3D-GS) is an effective paradigm for generalizable novel view synthesis. Given a set of calibrated images, recent methods directly predict an explicit Gaussian representation in a single forward pass~\cite{pixelsplat,mvsplat,depthsplat}, avoiding scene-specific optimization while retaining the efficient rendering properties of 3D-GS~\cite{3dgs}. 
However, as feed-forward reconstruction expands from sparse observations to longer sequences, both the reconstructed representation and the cross-view computation used to construct it become increasingly difficult to scale. A model that is efficient for a few input images may therefore become impractical when processing tens or hundreds of views.

% -------------------------------------------------------
\begin{figure}[!t]
    \centering
    \includegraphics[width=\linewidth]{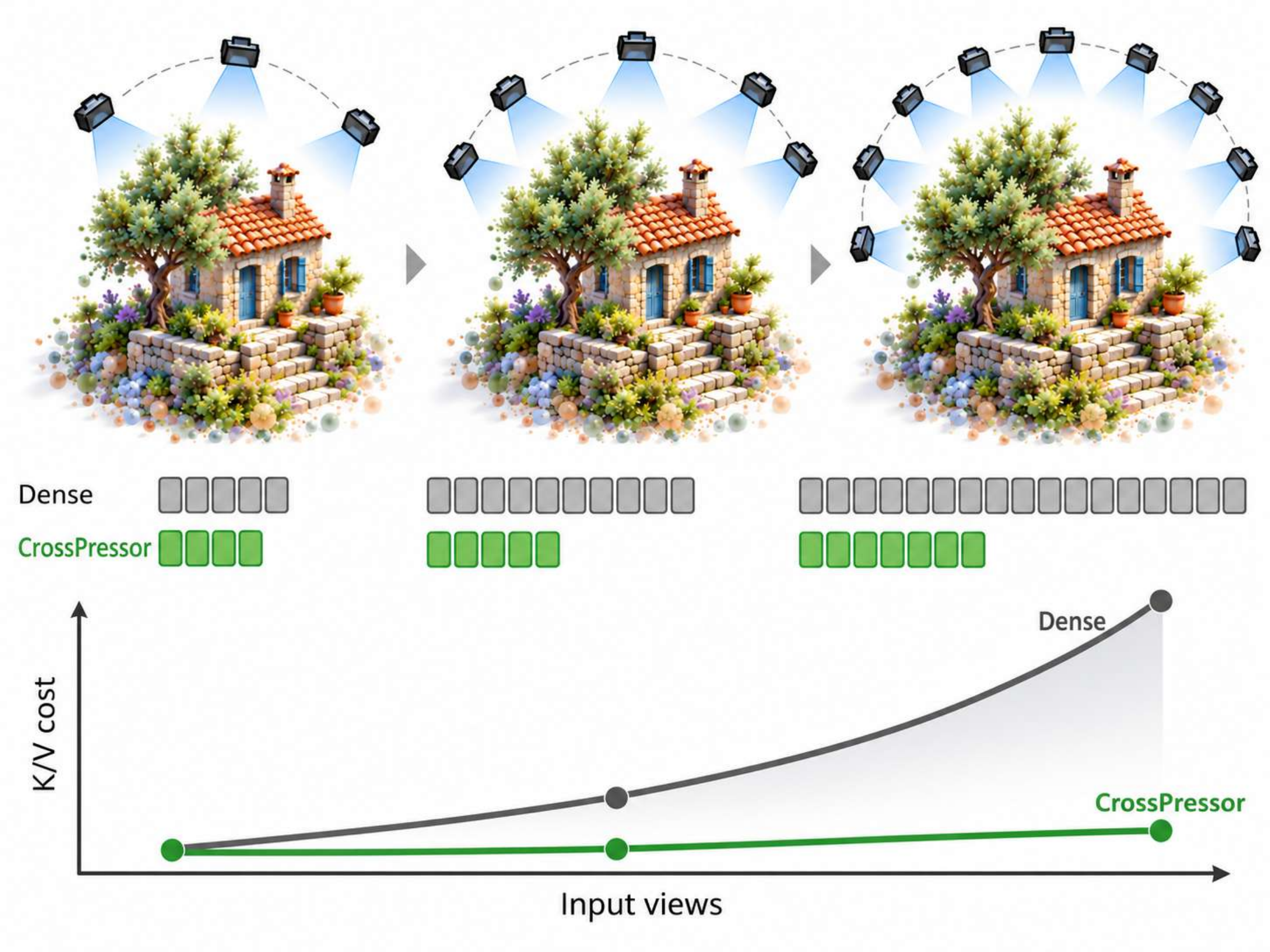}
    \caption{
    \textbf{Representation--fusion mismatch and \ours.}
    ZPressor~\cite{zpressor} bounds the decoder-facing representation by
    decoding Gaussians only from a compact set of anchor feature maps. As more context views are assigned to these anchors, the dense support features used to construct each anchor continue to grow. \ours keeps the anchor representation unchanged while routing a bounded support sequence across cameras and spatial locations, with spatial sampling offsets varied across fusion layers. Gray tokens denote dense support evidence, while green tokens indicate routed evidence from different support views.
    }
    \label{fig:teaser}
\end{figure}
% -------------------------------------------------------

Recent methods~\cite{freesplat,zpressor} have begun to address the issue of scaling in representation size. For instance, ZPressor~\cite{zpressor} partitions the input images into a fixed number of groups using camera poses. It selects one anchor view from each group and fuses the remaining support views into the corresponding anchor feature maps. Only these updated anchor features are passed to the pixel-aligned Gaussian decoder. Since the number of anchor views remains constant, adding additional support views to provide further scene information does not directly increase the input to the Gaussian decoder or the number of Gaussians.

This divide-and-conquer strategy, however, introduces a new scaling dilemma. Increasing the number of camera groups reduces the support burden per anchor, but also increases the number of decoder-facing anchors and generated Gaussians. Keeping the anchor set fixed instead preserves representation size while forcing each anchor to aggregate an increasing number of support views. This exposes a growing support-to-anchor fusion bottleneck even though the decoder-facing representation remains fixed.

We refer to this contradiction as the \emph{representation--fusion mismatch}: bounding the decoder-facing anchor representation does not bound the support evidence required to construct it. As more views are assigned to a fixed set of anchors, the representation remains compact while support-side K/V projection and cross-attention continue to grow.

Token reduction offers a direct way to reduce this cost. Learned methods adapt token selection to individual inputs and tasks~\cite{dynamicvit,spvit,tokencropr}, but typically require task-specific training or fine-tuning. When inserted into a feed-forward reconstruction pipeline, such modules may also require the host model to be jointly adapted, which is particularly costly for generalizable 3D-GS models trained on large-scale multi-view data. We therefore focus on inference-time reduction that can be applied directly to released checkpoints. Training-free methods~\cite{tome,optin,zerotprune} prune or merge redundant visual tokens, while Spark3R~\cite{spark3r} shows that K/V tokens can often tolerate stronger reduction than queries. In anchor-based fusion, however, the decoder-facing queries are already bounded by construction. The remaining challenge is to control the growing support-side K/V sequence while preserving evidence from all support views.

We therefore introduce \textbf{\ours}, a training-free method for scalable cross-view fusion in feed-forward 3D Gaussian Splatting. \ours constrains the growing support sequence through \emph{coverage-cycled evidence routing}: it assigns different spatial sampling offsets across support views and varies them across fusion layers, so additional views contribute evidence without proportionally increasing the K/V sequence. Routing is applied before K/V projection, and routed contexts are packed jointly across fusion layers. The original camera grouping, anchor representation, attention modules, and Gaussian decoder remain unchanged. Our main contributions are:
\begin{itemize}
    \item
    We identify the \emph{representation--fusion mismatch}: bounding the decoder-facing representation does not prevent support-to-anchor fusion from growing with the number of views.

    \item
    We propose \ours with \emph{coverage-cycled evidence routing}, which bounds the support-side sequence while distributing spatial evidence across support views and fusion layers.

    \item
    We route support features before K/V projection, reducing both support-side projection and attention cost without retraining. Experiments across multiple backbones and datasets demonstrate strong long-context acceleration while maintaining reconstruction quality.
\end{itemize}
\section{Related Work}
\label{sec:related}

\subsection{Feed-Forward 3D Gaussian Splatting}

3D Gaussian Splatting (3D-GS)~\cite{3dgs} represents a scene using explicit anisotropic Gaussian primitives and enables high-quality novel-view rendering through an efficient differentiable rasterizer.
While the standard approach~\cite{3dgs} relies on scene-by-scene optimization to generate a separate set of Gaussian parameter representations for each scene, feed-forward methods ~\cite{pixelsplat,mvsplat,mvsgaussian,c3gs,depthsplat,monosplat} directly predict scene representations from input images and thereby support novel-view synthesis without scene-specific optimization. This feed-forward paradigm is inspired by previous neural rendering methods~\cite{pixelnerf,ibrnet,mvsnerf}. These generalizable NeRF models are also capable of performing novel-view synthesis based on features extracted from input images, without requiring per-scene optimization.
PixelSplat~\cite{pixelsplat} predicts pixel-aligned Gaussians from sparse image pairs. MVSplat~\cite{mvsplat} further incorporates plane-sweep cost volumes and cross-view feature aggregation for geometric reconstruction, following the broader use of multi-view correspondence reasoning in learning-based MVS~\cite{mvsnet,casmvsnet,patchmatchnet,geomvsnet,icg-mvsnet,hu20223d}. DepthSplat~\cite{depthsplat} additionally integrates pretrained monocular depth features to strengthen geometric reasoning and cross-dataset generalization. This type of pixel-aligned method suffers from an inherent issue: the number of decoded Gaussians increases with the number of input viewpoints, as additional viewpoint-aligned feature maps lead to the prediction of more Gaussian primitives.

ZPressor~\cite{zpressor} reduces representation growth by compressing multiple input views into a compact set of decoder-facing anchor features. However, its support-to-anchor fusion still processes view-growing support evidence, so the cost of constructing the retained representation continues to increase with the number of input views. Our work is complementary to this representation compression. \ours keeps the original anchor representation and downstream Gaussian prediction pipeline unchanged, while making the support-to-anchor fusion more scalable.

\subsection{Visual Token Reduction}

Token reduction has been extensively studied for accelerating Vision Transformers~\cite{vit}. One class of methods learns input- or task-dependent token selection policies jointly with the model. For example, DynamicViT~\cite{dynamicvit} learns lightweight prediction modules for dynamic token sparsification, while more recent methods such as Token Cropr~\cite{tokencropr} learn task-aware token retention through auxiliary prediction objectives. Such learned reduction can adapt to the target task, but generally requires additional training or fine-tuning, and therefore does not directly preserve compatibility with existing reconstruction checkpoints.

Training-free methods instead modify token processing at inference time. Adaptive Token Sampling (ATS)~\cite{ats} performs parameter-free sampling using attention responses, but the attention information used for filtering must first be obtained.
ToMe~\cite{tome} merges similar tokens through bipartite matching, while Zero-TPrune~\cite{zerotprune} performs zero-shot token pruning without retraining. These methods provide general-purpose mechanisms for reducing visual tokens, but do not explicitly address how multi-view evidence should be distributed when many support cameras jointly construct a fixed decoder-facing representation.
Spark3R~\cite{spark3r} also studies training-free asymmetric token reduction for feed-forward 3D reconstruction methods~\cite{vggt,depthanything3} and observes that Key/Value tokens can often be reduced more aggressively than queries.
The scenario we address concerns a different type of bottleneck. In ZPressor-style anchor fusion, the decoder-facing anchor queries are already bounded by design, while the support evidence used to construct these queries continues to grow with the number of input views. Rather than selecting only a small subset of complete support views, \ours keeps every support camera represented and bounds the support-side K/V sequence by distributing preserved spatial evidence across support views and fusion layers. Furthermore, routing is applied to the raw support features before K/V projection, reducing both projection and cross-attention costs while remaining checkpoint-compatible and training-free.

% -------------------------------------------------------
\begin{figure*}[!t]
    \centering
    \includegraphics[width=\linewidth]{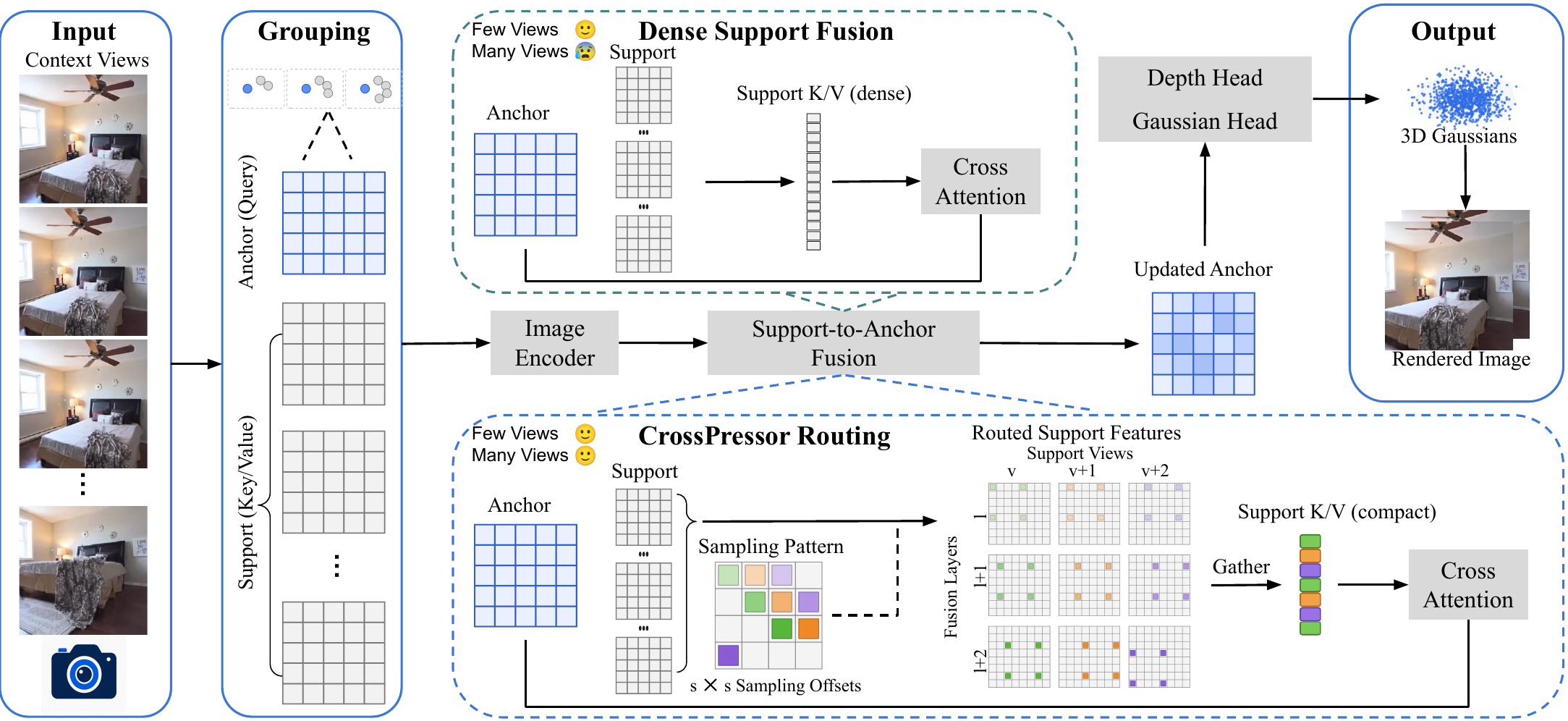}
    \caption{\textbf{Overview of \ours.} We keep the original ZPressor~\cite{zpressor} pipeline unchanged, including anchor selection, anchor--support grouping, the anchor-query pathway, and the downstream prediction modules. One representative anchor--support group is expanded to compare the fusion stage. The original dense fusion (top) uses all support features as K/V evidence for cross-attention, whereas \ours (bottom) preserves the full anchor-query pathway and routes only the support-side features before K/V projection. For cluster stride $s_c$, the $s_c\times s_c$ grid enumerates the $s_c^2$ possible spatial sampling offsets. We index the offset assigned to support view $v$ at fusion layer $\ell$ by $p_{c,v}^{\ell}=(r_c(v)+\kappa\ell)\bmod s_c^2$, where $r_c(v)$ is the support-view rank and $\kappa$ is the layer-wise increment. Moving across support views advances the sampling pattern by one offset, while moving across fusion layers advances it by $\kappa$, distributing the retained spatial evidence across both dimensions. The selected support features are gathered and packed into a compact sequence before the original K/V projection and cross-attention, producing the same updated-anchor interface for the unchanged downstream prediction pipeline.}
    \label{fig:pipeline}
\end{figure*}
% -------------------------------------------------------
\section{Method}
\label{sec:method}

Given posed images, the host reconstruction model first extracts view-aligned feature maps. ZPressor~\cite{zpressor}, as an inserted fusion module, groups these views around a fixed number of anchors and fuses support-view features into the corresponding anchor features, after which the host model continues with its original depth estimation and Gaussian parameter prediction.
\ours modifies only this support-to-anchor fusion stage without retraining. Instead of presenting the full support sequence to every fusion layer, it constructs a view-normalized routed sequence whose sampling offsets are distributed across support views and varied across fusion layers. Routing is performed before K/V projection, while the anchor queries, fusion blocks, downstream prediction modules, and output representation remain unchanged.

\subsection{Anchor-Based Fusion and Its Bottleneck}
\label{subsec:anchor_bottleneck}

We build on ZPressor~\cite{zpressor}, an anchor-based feature compression
module integrated into pixel-aligned feed-forward 3D-GS pipelines such as PixelSplat~\cite{pixelsplat}, MVSplat~\cite{mvsplat} and DepthSplat~\cite{depthsplat}. ZPressor~\cite{zpressor} selects $K$ anchor views using farthest-point sampling (FPS) over camera positions and assigns each remaining view to its nearest anchor, forming $K$ camera clusters.
For each cluster $c$, let $a_c$ denote the anchor-view index and $\mathcal S_c$ the ordered list of support views, with cluster size $m_c=1+|\mathcal S_c|$.
The support features are fused into the corresponding anchor feature, after which only the $K$ updated anchor maps continue through the original downstream reconstruction pipeline.

Let $\mathcal F_v\in\mathbb R^{H\times W\times C}$ denote the feature map of view $v$, where $H$ and $W$ are the image dimensions and $C$ is the feature dimension. With $T=HW$, we denote its spatially flattened token sequence by $\mathbf F_v\in\mathbb R^{T\times C}$. Thus, with $K$ fixed, additional support views can contribute multi-view evidence without directly increasing the number of decoder-facing feature maps or Gaussian predictions. We now isolate the support-to-anchor fusion. The initial anchor state $\mathbf Q_c^0\in\mathbb R^{T\times C}$ is

\begin{equation}
    \mathbf Q_c^0
    =
    \mathbf F_{a_c}.
    \label{eq:anchor_query}
\end{equation}

The support features are concatenated in their support-list order into
the dense cross-view memory $\mathbf M_c\in\mathbb R^{(m_c-1)T\times C}$:

\begin{equation}
    \mathbf M_c
    =
    \mathop{\operatorname{concat}}_{v\in\mathcal S_c}
    \mathbf F_v.
    \label{eq:dense_memory}
\end{equation}

Across the $L$ fusion layers, $\ell\in\{0,\ldots,L-1\}$,
the raw support memory $\mathbf M_c$ remains unchanged, whereas the anchor state $\mathbf Q_c^\ell\in\mathbb R^{T\times C}$ is progressively updated.
Let $\mathbf Y_c^\ell\in\mathbb R^{T\times C}$ denote the output of the dense cross-attention residual branch. Abstracting the layer-specific normalization and projection operations, we write

\begin{equation}
    \mathbf Y_c^\ell
    =
    \mathbf Q_c^\ell
    +
    \operatorname{CrossAttn}^{\ell}
    \left(
        \mathbf Q_c^\ell,
        \mathbf M_c
    \right).
    \label{eq:dense_fusion}
\end{equation}

The remaining operations of the original ZPressor block are unchanged
and produce the next anchor state $\mathbf Q_c^{\ell+1}$. The $T$ anchor queries therefore preserve the spatial extent of the decoder-facing representation, while the support sequence provides cross-view evidence used to refine it.

This design exposes a scaling tension. Increasing $K$ reduces the support burden per anchor but also increases the number of decoder-facing anchor maps and generated Gaussians. Keeping $K$ compact instead preserves representation size while forcing each anchor to try to aggregate more support views. For a cluster of $m_c$ views, the anchor retains $T$ queries whereas the dense support memory contains $(m_c-1)T$ tokens. Thus, bounding the decoded representation does not bound the support-side computation required to construct it. We call this inconsistency \emph{representation--fusion mismatch}.

\subsection{\ours}
\label{subsec:crosspressor}

\ours addresses this mismatch by leaving the decoder-facing anchor
representation unchanged and routing the view-growing support features
into a compact support sequence. Camera grouping, anchor selection, attention parameters, residual paths, and the downstream Gaussian predictor are unchanged. \ours therefore introduces no learnable parameters and can be applied to the corresponding pretrained checkpoint without retraining.

\paragraph{View-normalized support length.}
A fixed token count does not transfer well across feature resolutions.
We therefore express the routed support length relative to one dense
feature map:

\begin{equation}
    N_{KV}^{\max}
    =
    \beta_{KV}T,
    \label{eq:view_equiv_length}
\end{equation}

where $\beta_{KV}>0$ controls the target support length in dense-view
equivalents. We realize this target using a cluster-specific spatial stride
$s_c\in\mathcal D$, where $\mathcal D$ is a discrete set of
power-of-two strides $\{1,2,4,8,16,32,64\}$.
The routing stride $s_c$ is selected once for cluster $c$ and shared across all fusion layers; within the routing policy, only the spatial sampling offset varies with $\ell$. The stride policy chooses the smallest available $s_c$ satisfying

\begin{equation}
    m_c
    \left\lceil\frac{H}{s_c}\right\rceil
    \left\lceil\frac{W}{s_c}\right\rceil
    \leq
    N_{KV}^{\max}.
    \label{eq:stride_selection}
\end{equation}

The stride selection conservatively counts the full cluster of $m_c$
views, while the actual K/V sequence contains only the $m_c-1$ support
views.
As the number of support views grows, this rule selects a coarser stride,
reducing the spatial density retained from each view instead of letting
the support sequence grow proportionally.

\paragraph{Coverage-cycled evidence routing.}
The stride determines how many support tokens are retained, but not
which spatial locations they come from.
Selecting only a few complete support views reduces camera
participation, while applying the same sparse grid to every support view
and fusion layer repeatedly selects the same spatial locations.
We instead vary the spatial sampling pattern across both support views
and fusion layers.

After removing the anchor, support views retain the ascending order of
their original input-view indices.
Let $r_c(v)\in\{0,\ldots,|\mathcal S_c|-1\}$ denote the zero-based rank
of support view $v\in\mathcal S_c$.
For stride $s_c$, there are $s_c^2$ possible sampling offsets.
We index the offset assigned to support view $v$ at fusion layer $\ell$
by the sampling offset

\begin{equation}
    p_{c,v}^{\ell}
    =
    \bigl(
        r_c(v)+\kappa\ell
    \bigr)
    \bmod s_c^2,
    \label{eq:phase_index}
\end{equation}

where $p_{c,v}^{\ell}\in\{0,\ldots,s_c^2-1\}$ and $\kappa$ is a fixed
layer-wise increment. We use $L=6$ fusion layers and $\kappa=5$ (see supplementary~\cref{tab:phase_increment} for sensitivity).

The two terms in \cref{eq:phase_index} control variation across views
and layers, respectively.
Within a fusion layer, increasing the support rank by one advances the
sampling offset by one modulo $s_c^2$, so consecutive support views use
different spatial offsets.
For a fixed support view, moving to the next fusion layer advances the
offset by $\kappa$, changing the sampled spatial subset across layers.
Thus, support views cover different spatial patterns within a layer,
while each view contributes different locations across fusion layers.

We map the offset index to row and column offsets
$a_{c,v}^{\ell},b_{c,v}^{\ell}\in\{0,\ldots,s_c-1\}$:

\begin{align}
    a_{c,v}^{\ell}
    &=
    \left\lfloor
        \frac{p_{c,v}^{\ell}}{s_c}
    \right\rfloor,
    \\
    b_{c,v}^{\ell}
    &=
    p_{c,v}^{\ell}
    \bmod s_c.
    \label{eq:cycled_phase}
\end{align}

These offsets define the retained spatial locations

\begin{equation}
    \Omega_{c,v}^{\ell}
    =
    \left\{
        (i,j)
        \;\middle|\;
        \begin{array}{l}
        0\leq i<H,\;0\leq j<W,\\
        i\equiv a_{c,v}^{\ell}\!\!\!\pmod{s_c},\\
        j\equiv b_{c,v}^{\ell}\!\!\!\pmod{s_c}
        \end{array}
    \right\}.
    \label{eq:routing_set}
\end{equation}

For the evaluated feature layouts, the selected strides divide both
$H$ and $W$, so each sampling offset selects exactly $T/s_c^2$ spatial
locations.
The routed support length is therefore independent of the fusion layer:

\begin{equation}
    \widetilde N_c
    =
    |\mathcal S_c|
    \frac{T}{s_c^2}
    \leq
    N_{KV}^{\max}.
    \label{eq:length_constraint}
\end{equation}

Dense fusion instead uses
$N_c^{\mathrm{dense}}=|\mathcal S_c|T$ support tokens.
For these divisible layouts, the retained fraction within cluster $c$
is therefore

\begin{equation}
    \rho_c
    =
    \frac{\widetilde N_c}{N_c^{\mathrm{dense}}}
    =
    \frac{1}{s_c^2}.
    \label{eq:routed_fraction}
\end{equation}

This fraction is defined per cluster.
The QK retention reported in our experiments instead aggregates the
actual query--key interactions across clusters and scenes, whose strides
may differ.
The reduction is spatial rather than view-wise: every support camera
remains represented, with each contributing $T/s_c^2$ spatial tokens.

At layer $\ell$, the selected feature vectors are concatenated in
support-list order to form
$\widetilde{\mathbf M}_c^\ell
\in\mathbb R^{\widetilde N_c\times C}$:

\begin{equation}
    \widetilde{\mathbf M}_c^\ell
    =
    \mathop{\operatorname{concat}}_{v\in\mathcal S_c}
    \mathcal F_v
    \left[
        \Omega_{c,v}^{\ell}
    \right].
    \label{eq:routed_memory}
\end{equation}

Here,
$\mathcal F_v[\Omega]\in\mathbb R^{|\Omega|\times C}$
denotes the feature vectors at the selected spatial locations, stacked
in row-major order.
Within a layer, support views cycle through the $s_c^2$ possible
sampling phases, with phases repeating when
$|\mathcal S_c|>s_c^2$.
Across layers, the sampling offset of each support view is cyclically
shifted by $\kappa$ over the $s_c^2$ possible offsets, varying the
spatial evidence presented to the anchor without changing the routed
sequence length.
Setting $p_{c,v}^{\ell}=0$ for all support views and fusion layers gives
the Fixed variant used in our routing ablation.

\paragraph{Pre-projection routing.}
Routing is applied directly to the raw support features before the
layer-specific K/V normalization and projection.
Suppressing the internal attention-head decomposition, the projected
query has shape
$\widehat{\mathbf Q}_c^\ell\in\mathbb R^{T\times C}$,
while the routed keys and values satisfy
$\mathbf K_c^\ell,\mathbf V_c^\ell
\in\mathbb R^{\widetilde N_c\times C}$.
They are computed using the original learned operators:

\begin{align}
    \widehat{\mathbf Q}_c^\ell
    &=
    \operatorname{Proj}_{Q}^{\ell}
    \left(
        \operatorname{Norm}_{Q}^{\ell}
        \left(
            \mathbf Q_c^\ell
        \right)
    \right),
    \\
    \mathbf K_c^\ell,\mathbf V_c^\ell
    &=
    \operatorname{Proj}_{KV}^{\ell}
    \left(
        \operatorname{Norm}_{KV}^{\ell}
        \left(
            \widetilde{\mathbf M}_c^\ell
        \right)
    \right),
    \\
    \widetilde{\mathbf Y}_c^\ell
    &=
    \mathbf Q_c^\ell
    +
    \operatorname{Attn}^{\ell}
    \left(
        \widehat{\mathbf Q}_c^\ell,
        \mathbf K_c^\ell,
        \mathbf V_c^\ell
    \right).
    \label{eq:crosspressor_update}
\end{align}

The resulting residual output
$\widetilde{\mathbf Y}_c^\ell\in\mathbb R^{T\times C}$
is the routed counterpart of $\mathbf Y_c^\ell$ in
\cref{eq:dense_fusion}.
The remaining ZPressor operations are unchanged and produce
$\mathbf Q_c^{\ell+1}$ as in the dense model.
Because routing precedes K/V normalization and projection, discarded
support tokens avoid K/V normalization, projection, and attention
computation without requiring dense attention scores or learned
selection modules.

\subsection{Efficient Realization and Complexity}
\label{subsec:efficient_realization}

The anchor state changes across fusion layers, but the raw support
features $\{\mathcal F_v\}_{v\in\mathcal S_c}$ remain unchanged.
We therefore materialize all layer-specific routed contexts jointly
before entering the fusion-layer loop:

\begin{equation}
    \left\{
        \widetilde{\mathbf M}_c^\ell
    \right\}_{\ell=0}^{L-1}
    =
    \operatorname{Pack}
    \left(
        \left\{
            \mathcal F_v
        \right\}_{v\in\mathcal S_c},
        \left\{
            \Omega_{c,v}^{\ell}
        \right\}_{v,\ell}
    \right).
    \label{eq:multilayer_pack}
\end{equation}

Equivalently, the packed contexts can be viewed as a tensor in
$\mathbb R^{L\times\widetilde N_c\times C}$.
This prepacking step gathers the routed raw features for all layers,
while each layer still performs its own normalization, K/V projection,
and attention.
The optimized backend changes only this data movement: it preserves the
selected values and their ordering and does not reuse projected K/V
features across layers. The routing rule therefore changes the support dimension of the
cross-attention problem from $(m_c-1)T$ dense tokens to
$\widetilde N_c$ routed tokens, while leaving the $T$ anchor queries
unchanged.
Ignoring terms independent of the number of support views, the
support-dependent K/V projection and cross-attention cost of dense
fusion per layer is

\begin{equation}
    \mathcal O
    \left(
        (m_c-1)
        \left(
            TC^2
            +
            T^2C
        \right)
    \right).
    \label{eq:dense_complexity}
\end{equation}

For \ours, the corresponding support-dependent terms are

\begin{equation}
    \mathcal O
    \left(
        \widetilde N_c C^2
        +
        T\widetilde N_c C
    \right).
\end{equation}

Since
$\widetilde N_c\leq\beta_{KV}T$
in the evaluated layouts, these terms are bounded by

\begin{equation}
    \mathcal O
    \left(
        \beta_{KV}
        \left(
            TC^2
            +
            T^2C
        \right)
    \right).
    \label{eq:crosspressor_complexity}
\end{equation}

Thus, \ours removes support-view-count dependence from the dominant
layer-specific K/V projection and cross-attention terms.
The overall pipeline is not entirely independent of the number of input views: routing metadata still scales with the support count, and inherently per-view operations such as feature extraction remain unchanged.

\ours leaves the feature dimensionality, normalization and projection
layers, camera grouping, anchor representation, and Gaussian prediction
modules unchanged.
When $s_c=1$, every support token is retained in its original order,
exactly recovering the dense ZPressor computation and output.
The corresponding pretrained checkpoint can therefore be used without retraining.
\section{Experiments}
\label{sec:experiments}

\subsection{Experimental Setup}
\label{subsec:experimental_setup}

\begin{table*}[t]
    \centering
    \caption{
    \textbf{Training-free cross-view reduction at 256 context views on MVSplat--ZPressor.}
    All methods use the same pretrained checkpoint and preserve the decoder-facing anchor representation and \(393{,}216\) output Gaussians. Cross-view and E2E entries report latency and speedup relative to Exact. Speedups are computed from the underlying unrounded latency measurements.
    }
    \label{tab:main_comparison}
    \setlength{\tabcolsep}{4pt}
    \begin{tabular}{lccccccc}
        \toprule
        & \multicolumn{4}{c}{Rendering Quality}
        & \multicolumn{1}{c}{Attention Load}
        & \multicolumn{2}{c}{Efficiency} \\
        \cmidrule(lr){2-5}
        \cmidrule(lr){6-6}
        \cmidrule(lr){7-8}

        Method
        & PSNR $\uparrow$
        & SSIM $\uparrow$
        & LPIPS $\downarrow$
        & p5 $\Delta$PSNR $\uparrow$
        & QK Ret. $\downarrow$
        & \shortstack{Cross-view\\ms / speedup}
        & \shortstack{E2E\\ms / speedup} \\
        \midrule

        ZPressor~\cite{zpressor}
        & 23.1204
        & 0.81310
        & 0.18285
        & --
        & 100.00\%
        & 125.28 / 1.00$\times$
        & 432.16 / 1.00$\times$ \\

        + ToMe~\cite{tome}
        & 23.1205
        & 0.81310
        & 0.18284
        & -0.0032
        & 1.56\%
        & 317.50 / 0.40$\times$
        & 630.32 / 0.69$\times$ \\

        + Zero-TPrune~\cite{zerotprune}
        & 23.0966
        & 0.81282
        & 0.18306
        & -0.1666
        & 1.60\%
        & 6.91 / 18.12$\times$
        & 303.14 / 1.43$\times$ \\

        + \ours
        & 23.1203
        & 0.81311
        & 0.18284
        & -0.0063
        & 1.85\%
        & 4.80 / 26.12$\times$
        & 303.18 / 1.43$\times$ \\

        \bottomrule
    \end{tabular}
\end{table*}

\paragraph{Models and dataset.}
We evaluate \ours on the pretrained models released by ZPressor~\cite{zpressor}, where ZPressor is integrated into and jointly trained with several feed-forward Gaussian-splatting pipelines, including PixelSplat~\cite{pixelsplat}, MVSplat~\cite{mvsplat}, and DepthSplat~\cite{depthsplat}. We use the released MVSplat--ZPressor checkpoint as our primary model and additionally evaluate the method on the released DepthSplat--ZPressor checkpoint. Starting from these pretrained pipeline-specific checkpoints, we apply \ours directly at inference time and evaluate different variants without any retraining. For MVSplat--ZPressor, the model processes images at $256\times256$ resolution, partitions the context views into $6$ camera clusters, and therefore decodes the $6$ anchor feature maps into $393{,}216$ Gaussians. Unless otherwise stated, \ours uses \(\beta_{KV}=1.5\) with the coverage-cycled routing policy described in \cref{sec:method}.

We evaluate on the RealEstate10K (RE10K)~\cite{re10k} dataset using the official data loader and a paired long-context protocol with \(V\in\{36,64,128,192,256\}\) context views. The context sets are nested, and the same $8$ held-out target views are used across all context lengths.
We additionally evaluate cross-dataset generalization on ACID~\cite{acid} using the same model and settings.

\paragraph{Metrics and baselines.}
We report PSNR, SSIM~\cite{ssim}, and LPIPS~\cite{lpips} for novel-view synthesis, together with cross-view fusion latency, encoder latency, and end-to-end latency.
For each backbone, we use the corresponding released ZPressor model as the reference and denote its unmodified dense execution as \textbf{Exact}. Thus, Exact always refers to the original dense support cross-attention of the backbone-specific ZPressor checkpoint, without any inference-time token reduction.
We report p5 $\Delta$PSNR, the fifth percentile of the per-scene PSNR difference relative to Exact, to characterize lower-tail robustness, and QK Ret., the ratio of total retained to dense query--key interactions over the evaluation set.
All compared methods on MVSplat--ZPressor use the same released checkpoint without retraining and preserve the decoder-facing anchor representation and Gaussian count.

We compare against training-free token-reduction baselines on the same pretrained MVSplat--ZPressor model, including ToMe~\cite{tome} and Zero-TPrune~\cite{zerotprune}. We adapt ToMe~\cite{tome} to the support K/V stream while preserving all anchor queries, and reimplement Zero-TPrune~\cite{zerotprune} from the published method since no official code is available. For routing analysis, Fixed, View-only, Layer-only, and View+Layer use identical support views, K/V tokens, and QK interactions, differing only in sampling pattern assignment.

\paragraph{Runtime protocol.}
Runtime is measured on a single NVIDIA RTX~4090 (24G) with batch size one. Inputs are preloaded on the GPU, and data loading, metric computation, file I/O, compilation, and initial cache construction are excluded. For each evaluated backbone and context length, we run 5 fresh paired processes. Within each process, methods are measured in 20 alternating cycles after 10 warm-up iterations per mode, using CUDA events and explicit synchronization. We take the median over the 20 measurements within each process and then the median over the 5 process-level values.

\subsection{Long-Context Scaling and Quality}
\label{subsec:long_context}

\paragraph{Efficiency at 256 views.}
At the longest evaluated context, the low-interaction operating points exhibit different quality--efficiency trade-offs. ToMe~\cite{tome} retains only $1.56\%$ of the dense QK interactions and preserves rendering quality, but its measured cross-view latency increases to $317.50$ ms, exceeding Exact. Our Zero-TPrune~\cite{zerotprune} reimplementation converts a similarly small interaction ratio ($1.60\%$) into substantial acceleration, reducing cross-view latency to $6.91$ ms ($18.12\times$). However, its lower-tail quality degrades considerably, with p5 $\Delta$PSNR reaching $-0.1666$ dB. In comparison, \ours retains $1.85\%$ of dense QK interactions and achieves the lowest cross-view latency, $4.80$ ms ($26.12\times$), while maintaining near-exact rendering quality with only $-0.0063$ dB p5 $\Delta$PSNR. Zero-TPrune~\cite{zerotprune} and \ours reach essentially identical end-to-end latency, as the remaining runtime is dominated by components outside the modified fusion path. These results show that \ours combines low fusion latency with strong lower-tail robustness at a small retained interaction ratio.

\begin{figure}[!t]
    \centering
    \includegraphics[width=\linewidth]{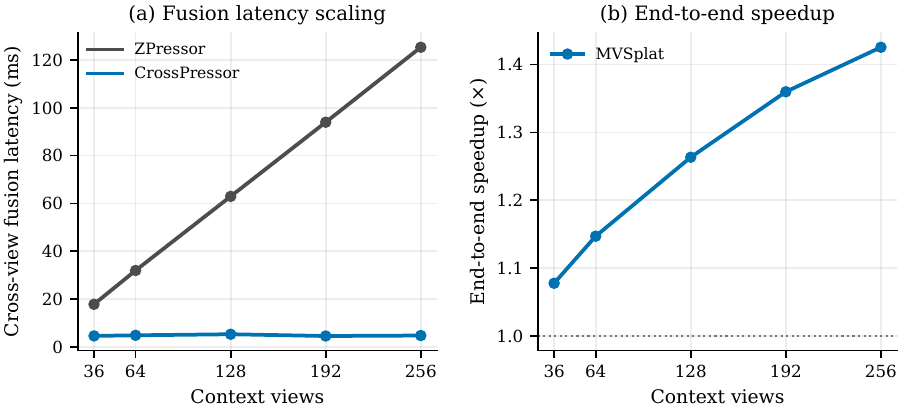}
    \caption{\textbf{Runtime scaling on MVSplat--ZPressor.}
    As the number of context views increases, dense cross-view fusion becomes progressively more expensive, whereas \ours remains nearly flat. The resulting end-to-end speedup reaches $1.43\times$ at 256 views.}
    \label{fig:runtime_scaling}
\end{figure}

\paragraph{Efficiency scaling.}
\cref{fig:runtime_scaling,tab:scaling_summary} show that the advantage
of \ours grows with context length.
As the number of views increases from 36 to 256, dense cross-view fusion becomes progressively more expensive, whereas the routed fusion latency remains nearly flat. Consequently, the cross-view speedup grows from $3.85\times$ to $26.12\times$. At $256$ views, \ours retains only $1.85\%$ of the dense QK interactions, confirming that the dominant view-growing attention workload is strongly reduced.
This fusion-level gain also translates to the full reconstruction pipeline in \cref{fig:runtime_scaling}: the end-to-end speedup increases from $1.08\times$ at 36 views to $1.43\times$ at 256 views. The smaller system-level gain is expected because substantial computation outside cross-view fusion remains unchanged. Overall, \ours keeps the dominant support-side K/V projection and cross-attention workload bounded, with measured fusion latency remaining nearly flat over the evaluated context range. Complete runtime measurements are provided in the supplementary.

\begin{figure}[!t]
    \centering
    \includegraphics[width=\linewidth]{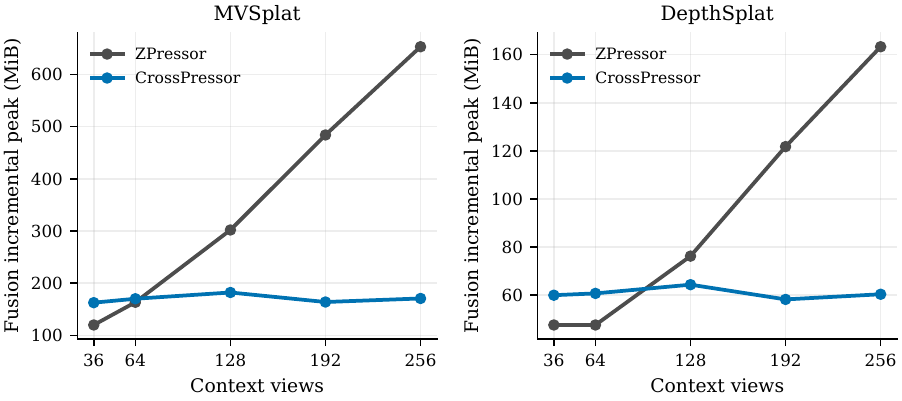}
    \caption{\textbf{Fusion-memory scaling on MVSplat--ZPressor and DepthSplat--ZPressor.} As the number of context views increases, dense fusion memory grows steadily, whereas the fusion-only incremental peak of \ours remains nearly flat. On MVSplat--ZPressor, the resulting reduction reaches $73.8\%$ at 256 views.}
    \label{fig:memory_scaling}
\end{figure}

\paragraph{Memory scaling.}
A similar trend appears in fusion memory (\cref{fig:memory_scaling,tab:scaling_summary}).
Routing and packing incur some additional memory at short contexts, but their cost changes little as more views are added. Dense fusion memory, in contrast, grows steadily with context length, and \ours reduces the fusion-only peak by $73.8\%$ at 256 views. The full-model peak changes little because most of the memory is used by components outside cross-view fusion. Detailed results for MVSplat~\cite{mvsplat} and DepthSplat~\cite{depthsplat} are provided in the supplementary material.

% -------------------------------------------------------
% -------------------------------------------------------

% -------------------------------------------------------
% -------------------------------------------------------

\begin{table}[t]
    \centering
    \caption{
    \textbf{Long-context scaling on MVSplat--ZPressor.}
    We report cross-view latency, its speedup over Exact, and the fusion-only incremental peak memory (Fusion mem.) as the number of context views increases. Cross-view latency and fusion memory are reported as Exact / \ours. Speedups are computed from the underlying unrounded latency measurements.
    }
    \label{tab:scaling_summary}
    \setlength{\tabcolsep}{4pt}
    \begin{tabular}{crrr}
        \toprule
        Views
        & Cross-view (ms)
        & \shortstack{Cross-view\\speedup $\uparrow$}
        & \shortstack{Fusion mem.\\(MiB) $\downarrow$} \\
        \midrule
        36 & 17.86 / 4.64 & $3.85\times$ & 120.0 / 162.8 \\
        64 & 31.98 / 4.90 & $6.53\times$ & 163.5 / 170.3 \\
        128 & 62.95 / 5.33 & $11.80\times$ & 302.0 / 182.3 \\
        192 & 93.95 / 4.60 & $20.43\times$ & 484.0 / 164.0 \\
        256 & 125.28 / 4.80 & $26.12\times$ & 653.0 / 170.8 \\
        \bottomrule
    \end{tabular}
\end{table}

\subsection{Routing and K/V Ablations}
\label{subsec:routing}

\begin{table}[t]
    \centering
    \caption{
    \textbf{Routing decomposition at 256 context views.}
    All variants use identical support views, K/V token counts, and QK interactions, differing only in the sampling phase. $N_{<-.05}$ counts scenes with $\Delta$PSNR below $-0.05$ dB.
    }
    \label{tab:routing_decomposition}
    \setlength{\tabcolsep}{2pt}
    \begin{tabular}{lrrr}
        \toprule
        Routing
        & p5 $\Delta$PSNR $\uparrow$
        & Worst $\Delta$PSNR $\uparrow$
        & $N_{<-.05}$ $\downarrow$ \\
        \midrule
        Fixed
        & -0.0204
        & -0.1055
        & 10 \\
        View-only
        & -0.0064
        & -0.0640
        & 1 \\
        Layer-only
        & -0.0154
        & -0.1071
        & 2 \\
        View+Layer
        & -0.0063
        & -0.0478
        & 0 \\
        \bottomrule
    \end{tabular}
\end{table}

We first isolate the effect of sampling offset assignment at 256 views.
All routing variants use identical support views, K/V token counts, and
QK interactions, differing only in how the sampling offset is assigned across views and fusion layers.
As shown in \cref{tab:routing_decomposition}, View-only provides most of the
lower-tail improvement, reducing the p5 degradation from $-0.0204$ to
$-0.0064$ dB, whereas Layer-only is less effective.
Combining both retains the View-only p5 behavior while improving the
worst case to $-0.0478$ dB and eliminating failures below $-0.05$ dB.
Thus, view-wise staggering is the main source of robustness, with
layer-wise cycling providing a complementary improvement in the
extreme tail.

\begin{table}[t]
    \centering
    \caption{
    \textbf{Sensitivity to the K/V target multiplier at 256 context views.} Results are evaluated on a fixed 200-scene subset. QK Ret. denotes the ratio of total retained to dense query--key interactions over this subset, and p5 $\Delta$PSNR is the fifth percentile of per-scene PSNR differences relative to Exact.
    }
    \label{tab:beta_sensitivity}
    \setlength{\tabcolsep}{8pt}
    \begin{tabular}{ccc}
        \toprule
        $\beta_{KV}$
        & QK Ret. $\downarrow$
        & p5 $\Delta$PSNR $\uparrow$ \\
        \midrule
        0.5 & 0.57\% & -0.0137 \\
        1.0 & 1.46\% & -0.0072 \\
        1.5 & 1.86\% & -0.0059 \\
        2.0 & 2.28\% & -0.0049 \\
        3.0 & 3.85\% & -0.0025 \\
        \bottomrule
    \end{tabular}
\end{table}

We next vary $\beta_{KV}$ to control the amount of retained support
evidence.
As shown in \cref{tab:beta_sensitivity}, larger values generally improve
lower-tail robustness while retaining more QK interactions.
Power-of-two strides may under-utilize the K/V budget, causing stepwise changes in QK retention as $\beta_{KV}$ varies.
We use $\beta_{KV}=1.5$ as the default trade-off between support-side
computation and robustness.

\subsection{Transfer Across Backbones and Datasets}
\label{subsec:transfer}

\begin{table}[t]
    \centering
    \caption{
    \textbf{Transfer to DepthSplat--ZPressor.}
    We apply \ours without retraining. Quality changes are relative to Exact over the same $1{,}034$ scenes; cross-view latency is reported as Exact / \ours in milliseconds.
    }
    \label{tab:depthsplat_transfer}
    \setlength{\tabcolsep}{4pt}
    \begin{tabular}{crrr}
        \toprule
        Views
        & p5 $\Delta$PSNR $\uparrow$
        & Cross-view (ms)
        & \shortstack{Cross-view\\speedup $\uparrow$} \\
        \midrule
        36
        & -0.0052
        & 5.76 / 5.44
        & $1.06\times$ \\
        64
        & -0.0060
        & 6.49 / 5.52
        & $1.18\times$ \\
        128
        & -0.0110
        & 8.77 / 5.58
        & $1.57\times$ \\
        192
        & -0.0087
        & 12.67 / 5.64
        & $2.25\times$ \\
        256
        & -0.0065
        & 16.40 / 5.55
        & $2.95\times$ \\
        \bottomrule
    \end{tabular}
\end{table}

We further apply \ours to DepthSplat--ZPressor~\cite{depthsplat,zpressor} without retraining. As shown in \cref{tab:depthsplat_transfer}, the p5 $\Delta$PSNR remains above $-0.011$ dB across all context lengths.
The long-context scaling behavior transfers as well: routed cross-view latency remains nearly flat, while the dense counterpart grows with the number of views, increasing the cross-view speedup from $1.06\times$ to $2.95\times$.

The smaller system-level gain on DepthSplat~\cite{depthsplat} reflects its runtime composition. MVSplat~\cite{mvsplat} performs ZPressor~\cite{zpressor} fusion on $64\times64$ features, whereas DepthSplat~\cite{depthsplat} uses a $32\times32$ support layout; at 256 views, dense cross-view fusion accounts for $29.0\%$ and $7.1\%$ of their end-to-end latency, respectively. Consequently, the end-to-end speedup is smaller, and the full-model memory peak remains nearly unchanged, as both are dominated by components outside the modified fusion stage (~\cref{tab:supp_runtime_scaling}).

We also evaluate the same MVSplat--ZPressor checkpoint on ACID~\cite{acid} dataset without adaptation. Over 599 scenes at 256 context views, \ours remains close to Exact, with a p5 $\Delta$PSNR of $-0.0079$ dB and a worst-case degradation of $-0.0410$ dB. Its cross-view latency decreases from $125.15$ ms to $5.02$ ms ($24.91\times$), with a $1.43\times$ end-to-end speedup. These results show a similar quality--efficiency trade-off on a different image distribution. Detailed results are provided in the supplementary material.
\section{Conclusion}
\label{sec:conclusion}

We propose \ours for scalable cross-view fusion in feed-forward 3D-GS. By identifying the representation--fusion mismatch, we show that bounding the decoder-facing anchor representation does not prevent the support-side fusion cost from growing with the number of views.
\ours addresses this issue through training-free coverage-cycled evidence routing while leaving the anchor representation and downstream prediction pipeline unchanged. Across long-context evaluations, it substantially reduces cross-view computation and fusion memory while maintaining rendering quality. These results suggest that scalable feed-forward reconstruction requires controlling not only the final representation size, but also the cost and coverage of the multi-view evidence used to construct it.
%%%%%%%%% 

%%%%%%%%% Acknowledgements
\vspace{4pt}
\noindent\textbf{Acknowledgements}
We appreciate the Austrian Scientific Computing (ASC) infrastructure for computing resources.

%%%%%%%%% REFERENCES
{
    \small
    \bibliographystyle{ieeenat_fullname}
    \bibliography{main}
}
%%%%%%%%% 

%%%%%%%%% SUPP
\clearpage

This supplementary material complements the main paper with additional
implementation details, evaluation protocols, and experimental
results.

\section{Implementation and Evaluation Details}
\label{sec:supp_implementation}

\subsection{Implementation Details}

\paragraph{Feature interface and routing configuration.}
We apply \ours at the ZPressor~\cite{zpressor} fusion interface without modifying the pretrained host backbone~\cite{mvsplat,depthsplat} or downstream Gaussian prediction modules. For MVSplat--ZPressor, each view provides a $64\times64$ feature map with $C=128$, giving $T=4096$ tokens per view.
For DepthSplat--ZPressor, the corresponding feature map is $32\times32$ with $C=128$ and $T=1024$.
Both integrations use $6$ anchor clusters and $6$ fusion layers. Unless otherwise stated, we use $\beta_{KV}=1.5$ and $\kappa=5$.

Anchor selection follows the original ZPressor~\cite{zpressor} implementation: FPS is performed on camera positions, followed by nearest-anchor assignment. After removing the anchor from each cluster, support views retain the ascending order of their original input-view indices; this order defines the zero-based rank $r_c(v)$ used by the routing rule.
The spatial stride $s_c$ is selected once per cluster from $\{1,2,4,8,16,32,64\}$ and is shared by all six fusion layers, while the sampling pattern, which indexes the spatial offset, varies across layers.
For the $64\times64$ and $32\times32$ feature grids used in our experiments, the selected power-of-two strides divide the spatial dimensions exactly, so each routed phase contains $T/s_c^2$ tokens per support view.

\paragraph{Pre-projection packing.}
Routing is applied to the raw support features before the layer-specific K/V normalization and projection. Because the raw support memory is shared across the six fusion layers, the optimized implementation constructs the routed indices for all layers and jointly gathers the corresponding contexts into a layer-major packed representation equivalent to $[L,\widetilde N_c,C]$.
Each fusion layer then independently applies its original normalization, K/V projection, and attention operation; projected K/V features are not shared across layers.
Within a packed context, tokens follow support-view order and row-major spatial order within each selected phase.

We additionally verify the optimized packing implementation against the reference routing path. A 36-view equivalence test gives zero maximum difference between the reference and optimized packed contexts, with identical intermediate anchor states, Gaussian parameters, and rendered outputs. At $s_c=1$, the packed context recovers the complete ordered support sequence and the resulting output matches Exact.

\subsection{Evaluation Protocol}

\paragraph{Rendering quality.}
The main quality evaluation uses $1{,}034$ RealEstate10K (RE10K)~\cite{re10k} scenes with 256 context views and 8 target views per scene, for a total of $8{,}272$ target renders.
The same evaluation scope is used for the MVSplat--ZPressor and DepthSplat--ZPressor quality experiments.
PSNR, SSIM, and LPIPS are first computed for individual target views, then averaged over the 8 targets within each scene, and finally averaged over scenes for the aggregate metrics.

For paired quality statistics, we first compute the scene-level
difference
\begin{equation}
    \Delta_i
    =
    \overline{\mathrm{PSNR}}_{i,\mathrm{method}}
    -
    \overline{\mathrm{PSNR}}_{i,\mathrm{Exact}},
\end{equation}
where the bar denotes the mean over the 8 target views of scene $i$.
Mean, fifth-percentile (p5), worst-case $\Delta$PSNR, and the reported degradation counts are computed over these scene-level differences.
The $\beta_{KV}$ and sampling offsets sensitivity studies use the same fixed 200-scene subset sampled uniformly over the frozen $1{,}034$-scene evaluation order.

\paragraph{QK interaction accounting.}
QK Ret. measures the fraction of dense query--key interactions retained over the evaluation scope:
\begin{equation}
    \mathrm{QKRet.}
    =
    \frac{
        \sum_{i,c,\ell}
        N_{Q,ic\ell}
        N^{\mathrm{routed}}_{K,ic\ell}
    }{
        \sum_{i,c,\ell}
        N_{Q,ic\ell}
        N^{\mathrm{dense}}_{K,ic\ell}
    },
    \qquad
    N_{Q,ic\ell}=T.
    \label{eq:supp_qk_retention}
\end{equation}
The ratio is accumulated globally over the scenes, clusters, and fusion layers included in each experiment.
Consequently, values reported for the full evaluation set, the fixed 200-scene subset, and the runtime-scaling benchmark correspond to their respective evaluation scopes.

\paragraph{Runtime scaling.}
The long-context runtime experiments use $V\in\{36,64,128,192,256\}$ context views.
The context sets are nested according to a consistent view ordering, and the target views remain unchanged across context lengths. Exact and \ours use identical context and target indices at each setting.
The DepthSplat~\cite{depthsplat} runtime experiment follows the same scaling protocol.

\paragraph{Memory measurement.}
We report allocated GPU memory rather than allocator-reserved memory.
For the fusion-only measurement, we record the live allocated memory immediately before fusion, reset the CUDA peak-memory statistics, and measure the maximum allocated memory during synchronized fusion execution.
The reported fusion-only incremental peak is the difference between this peak and the live allocation immediately before fusion, and therefore includes the routing and packing workspace used by \ours.

For the full-model measurement, peak-memory statistics are reset after warm-up and the maximum allocated memory is recorded over the complete forward execution.
Memory values are converted from bytes to MiB using $2^{20}$ bytes per MiB. Reported memory reductions are computed from the underlying unrounded measurements.

\subsection{Baseline and Ablation Settings}

\paragraph{Baseline configurations.}
For each backbone, we start from the corresponding officially released ZPressor~\cite{zpressor} checkpoint. All compared reduction methods for that backbone use the same released checkpoint and leave the learned ZPressor weights and downstream Gaussian prediction modules unchanged.
For each backbone, ``Exact'' denotes the unmodified dense execution of the corresponding released ZPressor checkpoint.

Our ToMe~\cite{tome} adaptation preserves the anchor-query pathway and applies token merging to the projected support K/V sequence. The operating point used in the main comparison retains $1.56\%$ of the dense QK interactions.
For Zero-TPrune~\cite{zerotprune}, we adapt its token-selection rule to the ZPressor support path: anchor queries remain unchanged, while complete support views are selected before the standard K/V projection. The operating point used in the main comparison retains $1.60\%$ of dense QK interactions.
Neither baseline requires additional training.
\ours uses $\beta_{KV}=1.5$, View+Layer routing, and $\kappa=5$, also without retraining.

\paragraph{Routing decomposition.}
The routing ablations differ only in how the sampling pattern is assigned. For support rank $r_c(v)$ and fusion layer $\ell$, we use
\begin{align}
    \text{Fixed:}\qquad
    p_{c,v}^{\ell}
    &= 0,
    \\
    \text{View-only:}\qquad
    p_{c,v}^{\ell}
    &=
    r_c(v)\bmod s_c^2,
    \\
    \text{Layer-only:}\qquad
    p_{c,v}^{\ell}
    &=
    \kappa\ell\bmod s_c^2,
    \\
    \text{View+Layer:}\qquad
    p_{c,v}^{\ell}
    &=
    \bigl(r_c(v)+\kappa\ell\bigr)\bmod s_c^2 .
\end{align}
We use $\kappa=5$ whenever layer-wise cycling is enabled. Across these variants, the support assignments, cluster strides, routed token counts, and total QK interactions are unchanged; only the offset differs.

In \cref{tab:supp_routing_decomposition}, we additionally report two phase-coverage statistics to characterize the routing patterns.
For each scene--cluster--layer block, View Cov. measures the fraction of the $s_c^2$ sampling offsets represented across its support views, averaged equally over all such blocks.
Layer Cov. measures, for each support view, the fraction of sampling offsets visited across the six fusion layers, averaged equally over support views.
These two quantities respectively characterize offset diversity across support cameras and across fusion layers.

\begin{table*}[!t]
    \centering
    \caption{
    \textbf{Detailed rendering-quality comparison at 256 context views on RE10K~\cite{re10k}.}
    All methods are evaluated on the same \(1{,}034\) RE10K scenes and
    \(8{,}272\) paired target views.
    Paired PSNR statistics are computed scene-wise relative to Exact.
    Failures report the number of scenes with
    \(\Delta\mathrm{PSNR}<-0.01/-0.03/-0.05\) dB, respectively.
    }
    \label{tab:supp_quality_comparison}
    \begin{tabular}{lccccccc}
        \toprule
        & \multicolumn{3}{c}{Rendering Quality}
        & \multicolumn{4}{c}{Paired PSNR Robustness} \\
        \cmidrule(lr){2-4}
        \cmidrule(lr){5-8}
        Method
        & PSNR $\uparrow$
        & SSIM $\uparrow$
        & LPIPS $\downarrow$
        & Mean $\Delta$
        & p5 $\Delta$
        & Worst $\Delta$
        & \shortstack{Failures\\(.01/.03/.05)} \\
        \midrule

        ZPressor~\cite{zpressor}
        & 23.12037
        & 0.813096
        & 0.182849
        & --
        & --
        & --
        & -- \\

        + ToMe~\cite{tome}
        & 23.12052
        & 0.813099
        & 0.182839
        & +0.0002
        & -0.0032
        & -0.1041
        & 6 / 1 / 1 \\

        + Zero-TPrune~\cite{zerotprune}
        & 23.09657
        & 0.812823
        & 0.183055
        & -0.0238
        & -0.1666
        & -1.5052
        & 536 / 418 / 334\\

        + \ours
        & 23.12029
        & 0.813105
        & 0.182836
        & -0.0001
        & -0.0063
        & -0.0478
        & 21 / 2 / 0 \\

        \bottomrule
    \end{tabular}
\end{table*}
\begin{table*}[!t]
    \centering
    \caption{
    \textbf{Detailed runtime breakdown at 256 context views on RE10K~\cite{re10k}.}
    We report latency at four nested execution scopes: cross-view fusion,
the complete ZPressor module, the encoder, and the end-to-end pipeline.
Cross-view entries reproduce the fusion-region measurements used in the
main comparison.
Speedups are computed from the underlying unrounded latency measurements.
    }
    \label{tab:supp_runtime_breakdown}
    \setlength{\tabcolsep}{4.2pt}
    \begin{tabular}{lcccccccc}
        \toprule
        & \multicolumn{4}{c}{Latency (ms)}
        & \multicolumn{4}{c}{Speedup over Exact} \\
        \cmidrule(lr){2-5}
        \cmidrule(lr){6-9}
        Method
        & Cross-view $\downarrow$
        & ZPressor $\downarrow$
        & Encoder $\downarrow$
        & E2E $\downarrow$
        & Cross-view $\uparrow$
        & ZPressor $\uparrow$
        & Encoder $\uparrow$
        & E2E $\uparrow$ \\
        \midrule
        ZPressor~\cite{zpressor}
        & 125.28
        & 142.80
        & 418.59
        & 432.16
        & 1.00$\times$
        & 1.00$\times$
        & 1.00$\times$
        & 1.00$\times$ \\

        + ToMe~\cite{tome}
        & 317.50
        & 342.19
        & 616.68
        & 630.32
        & 0.40$\times$
        & 0.42$\times$
        & 0.68$\times$
        & 0.69$\times$ \\

        + Zero-TPrune~\cite{zerotprune}
        & 6.91
        & 14.21
        & 289.49
        & 303.14
        & 18.12$\times$
        & 10.05$\times$
        & 1.45$\times$
        & 1.43$\times$ \\

        + \ours
        & 4.80
        & 14.12
        & 289.64
        & 303.18
        & 26.12$\times$
        & 10.11$\times$
        & 1.45$\times$
        & 1.43$\times$ \\
        \bottomrule
    \end{tabular}
\end{table*}

\begin{table*}[!t]
    \centering
    \caption{
    \textbf{Detailed rendering-quality comparison at 256 context views on ACID~\cite{acid}.}
    All methods are evaluated on the same \(599\) ACID scenes and
    \(4{,}792\) paired target views.
    Paired PSNR statistics are computed scene-wise relative to Exact.
    Failures report the number of scenes with
    \(\Delta\mathrm{PSNR}<-0.01/-0.03/-0.05\) dB, respectively.
    }
    \label{tab:supp_quality_comparison_acid}
    \begin{tabular}{lccccccc}
        \toprule
        & \multicolumn{3}{c}{Rendering Quality}
        & \multicolumn{4}{c}{Paired PSNR Robustness} \\
        \cmidrule(lr){2-4}
        \cmidrule(lr){5-8}
        Method
        & PSNR $\uparrow$
        & SSIM $\uparrow$
        & LPIPS $\downarrow$
        & Mean $\Delta$
        & p5 $\Delta$
        & Worst $\Delta$
        & \shortstack{Failures\\(.01/.03/.05)} \\
        \midrule

        ZPressor~\cite{zpressor}
        & 24.63131
        & 0.763763
        & 0.213089
        & --
        & --
        & --
        & -- \\

        + ToMe~\cite{tome}
        & 24.63146
        & 0.763766
        & 0.213085
        & +0.0002
        & -0.0039
        & -0.0204
        & 3 / 0 / 0 \\

        + Zero-TPrune~\cite{zerotprune}
        & 24.60262
        & 0.763369
        & 0.213654
        & -0.0287
        & -0.2983
        & -0.7663
        & 309 / 253 / 205 \\

        + \ours
        & 24.63128
        & 0.763764
        & 0.213084
        & 0.0000
        & -0.0079
        & -0.0410
        & 17 / 3 / 0 \\

        \bottomrule
    \end{tabular}
\end{table*}
\begin{table*}[!t]
    \centering
    \caption{
    \textbf{Detailed runtime breakdown at 256 context views on ACID~\cite{acid}.}
    We report latency at four nested execution scopes: cross-view fusion,
    the complete ZPressor module, the encoder, and the end-to-end pipeline.
    Speedups are computed from the underlying unrounded latency measurements
    relative to the same dense ZPressor execution.
    }
    \label{tab:supp_runtime_breakdown_acid}
    \setlength{\tabcolsep}{4.2pt}
    \begin{tabular}{lcccccccc}
        \toprule
        & \multicolumn{4}{c}{Latency (ms)}
        & \multicolumn{4}{c}{Speedup over Exact} \\
        \cmidrule(lr){2-5}
        \cmidrule(lr){6-9}
        Method
        & Cross-view $\downarrow$
        & ZPressor $\downarrow$
        & Encoder $\downarrow$
        & E2E $\downarrow$
        & Cross-view $\uparrow$
        & ZPressor $\uparrow$
        & Encoder $\uparrow$
        & E2E $\uparrow$ \\
        \midrule

        ZPressor~\cite{zpressor}
        & 125.15
        & 142.77
        & 417.66
        & 427.78
        & 1.00$\times$
        & 1.00$\times$
        & 1.00$\times$
        & 1.00$\times$ \\

        + ToMe~\cite{tome}
        & 319.14
        & 343.74
        & 617.77
        & 627.92
        & 0.39$\times$
        & 0.42$\times$
        & 0.68$\times$
        & 0.68$\times$ \\

        + Zero-TPrune~\cite{zerotprune}
        & 6.96
        & 14.27
        & 289.59
        & 299.64
        & 17.98$\times$
        & 10.01$\times$
        & 1.44$\times$
        & 1.43$\times$ \\

        + \ours
        & 5.02
        & 14.30
        & 289.11
        & 299.18
        & 24.91$\times$
        & 9.98$\times$
        & 1.44$\times$
        & 1.43$\times$ \\

        \bottomrule
    \end{tabular}
\end{table*}

\section{Additional Experimental Results}

\subsection{Detailed Quality and Runtime Results}

\subsubsection{Rendering-Quality Results}

We provide a more detailed rendering-quality evaluation at 256 context views in \cref{tab:supp_quality_comparison}, as a supplement to the contents of \cref{tab:main_comparison}.
In addition to the PSNR, SSIM, and LPIPS metrics reported in the main paper, we report paired scene-level quality changes relative to Exact, including mean, lower-tail, and worst-case $\Delta$PSNR together with the number of scenes falling below predefined degradation thresholds.
These statistics distinguish reduction methods that achieve similar average rendering quality but differ in their robustness on difficult scenes.

The paired statistics reveal different quality--efficiency trade-offs among the training-free reduction methods.
ToMe~\cite{tome} preserves rendering quality while retaining only a
small fraction of the dense QK interactions, but its measured
cross-view latency remains higher than Exact.
Zero-TPrune~\cite{zerotprune} instead achieves substantial fusion-level speedup at a similarly low interaction ratio, but exhibits a pronounced lower tail: a small subset of scenes suffers considerably larger degradation despite competitive average quality. Together, these results show that effective long-context reduction must be both computationally inexpensive and robust in how it preserves multi-view evidence.

\ours provides this combination through structured pre-projection routing.
It turns the reduced support sequence into actual wall-clock acceleration without content-dependent token matching, while distributing the retained spatial evidence across support views to avoid severe lower-tail failures.
Consequently, both aggregate rendering quality and per-scene degradation remain close to Exact. This motivates evaluating reduction methods jointly by interaction count, measured runtime, and lower-tail quality rather than by QK reduction alone.

\subsubsection{Runtime Breakdown}

We provide the complete runtime breakdown at 256 context views in
\cref{tab:supp_runtime_breakdown}.
For each method, we report latency and speedup at four nested scopes: cross-view fusion, the complete ZPressor~\cite{zpressor} module, the encoder, and the end-to-end reconstruction pipeline. These measurements show how acceleration within cross-view fusion propagates through the surrounding reconstruction system.

The difference between fusion-level and end-to-end speedups reflects the
amount of computation left unchanged by the reduction.
The measurements also show that a small retained QK interaction ratio
does not by itself imply low wall-clock fusion latency.
We therefore report both QK retention and measured cross-view latency
rather than using interaction count alone as a proxy for efficiency.
Even large fusion-level savings translate to a more modest end-to-end
gain once feature extraction, geometry estimation, and Gaussian
prediction are included.

\subsection{Cross-Dataset Evaluation on ACID}

We additionally evaluate the same pretrained MVSplat--ZPressor model on
ACID~\cite{acid}.
We retain 599 scenes satisfying the 256-context-view protocol, yielding 4,792 paired target views.
All methods use the same scene list and identical context and target indices.

As shown in \cref{tab:supp_quality_comparison_acid},
the quality trends observed on RE10K~\cite{re10k} persist on ACID~\cite{acid}.
Both ToMe~\cite{tome} and \ours preserve the aggregate rendering quality of Exact.
\ours maintains a bounded lower tail, with a worst-case
$\Delta$PSNR of $-0.0410$ dB and no scene degrading by more than
$0.05$ dB.
In contrast, Zero-TPrune~\cite{zerotprune} exhibits substantially larger
degradation, with a p5 $\Delta$PSNR of $-0.2983$ dB and 205 of 599 scenes
falling below $-0.05$ dB.

The runtime ordering is also consistent with RE10K~\cite{re10k} (\cref{tab:supp_runtime_breakdown_acid}).
ToMe~\cite{tome} remains slower than dense ZPressor despite aggressive interaction
reduction, while Zero-TPrune and \ours both achieve about $1.43\times$ end-to-end acceleration.
At the cross-view level, \ours reduces latency from $125.15$ to
$5.02$ ms, compared with $6.96$ ms for Zero-TPrune~\cite{zerotprune}.
Overall, ACID~\cite{acid} shows the same quality--efficiency pattern observed on RE10K~\cite{re10k}: ToMe~\cite{tome} preserves quality but remains slower than Exact, Zero-TPrune~\cite{zerotprune} is faster but less robust, whereas \ours maintains near-Exact rendering quality with low cross-view latency.

\begin{table*}[!t]
    \centering
    \caption{
    \textbf{Detailed runtime scaling on MVSplat--ZPressor and DepthSplat--ZPressor.}
    We report latency for cross-view fusion, the complete ZPressor module,
    encoder, and end-to-end reconstruction at increasing context lengths.
    For each backbone, ``Exact'' denotes the unmodified dense execution of the corresponding released ZPressor model. All latency measurements are in milliseconds. Speedups are computed from the underlying unrounded latency measurements.
    }
    \label{tab:supp_runtime_scaling}
    \setlength{\tabcolsep}{3.0pt}
    \small
    \begin{tabular}{l r
        rr r
        rr r
        rr r
        rr r}
        \toprule
        & 
        & \multicolumn{3}{c}{Cross-view}
        & \multicolumn{3}{c}{ZPressor}
        & \multicolumn{3}{c}{Encoder}
        & \multicolumn{3}{c}{End-to-end (E2E)} \\
        \cmidrule(lr){3-5}
        \cmidrule(lr){6-8}
        \cmidrule(lr){9-11}
        \cmidrule(lr){12-14}
        Backbone
        & Views
        & Exact & Ours & Speedup
        & Exact & Ours & Speedup
        & Exact & Ours & Speedup
        & Exact & Ours & Speedup \\
        \midrule

        \multirow{5}{*}{MVSplat}
        & 36
        & 17.86 & 4.64 & $3.85\times$
        & 27.32 & 13.29 & $2.05\times$
        & 178.09 & 164.24 & $1.08\times$
        & 191.97 & 178.12 & $1.08\times$ \\

        & 64
        & 31.98 & 4.90 & $6.53\times$
        & 42.33 & 13.70 & $3.09\times$
        & 210.05 & 181.39 & $1.16\times$
        & 224.01 & 195.29 & $1.15\times$ \\

        & 128
        & 62.95 & 5.33 & $11.80\times$
        & 75.60 & 14.29 & $5.29\times$
        & 279.21 & 218.33 & $1.28\times$
        & 291.33 & 230.59 & $1.26\times$ \\

        & 192
        & 93.95 & 4.60 & $20.43\times$
        & 108.98 & 13.65 & $7.98\times$
        & 348.56 & 252.96 & $1.38\times$
        & 361.05 & 265.51 & $1.36\times$ \\

        & 256
        & 125.28 & 4.80 & $26.12\times$
        & 142.80 & 14.12 & $10.11\times$
        & 418.59 & 289.64 & $1.45\times$
        & 432.16 & 303.18 & $1.43\times$ \\

        \midrule

        \multirow{5}{*}{DepthSplat}
        & 36
        & 5.76 & 5.44 & $1.06\times$
        & 11.08 & 10.50 & $1.06\times$
        & 101.45 & 100.84 & $1.01\times$
        & 118.33 & 117.74 & $1.01\times$ \\

        & 64
        & 6.49 & 5.52 & $1.18\times$
        & 11.19 & 10.60 & $1.05\times$
        & 114.28 & 113.80 & $1.00\times$
        & 131.19 & 130.71 & $1.00\times$ \\

        & 128
        & 8.77 & 5.58 & $1.57\times$
        & 12.88 & 10.70 & $1.20\times$
        & 145.76 & 143.63 & $1.01\times$
        & 161.11 & 158.95 & $1.01\times$ \\

        & 192
        & 12.67 & 5.64 & $2.25\times$
        & 17.28 & 10.93 & $1.58\times$
        & 179.94 & 173.53 & $1.04\times$
        & 195.91 & 189.48 & $1.03\times$ \\

        & 256
        & 16.40 & 5.55 & $2.95\times$
        & 21.57 & 10.89 & $1.98\times$
        & 213.56 & 202.79 & $1.05\times$
        & 230.38 & 219.60 & $1.05\times$ \\

        \bottomrule
    \end{tabular}
\end{table*}
\begin{table*}[!t]
    \centering
    \caption{
    \textbf{Detailed memory scaling on MVSplat--ZPressor and DepthSplat--ZPressor.}
    We report the fusion-only incremental peak memory used in the scaling analysis,
    together with the full-model peak allocated memory.
    For each backbone, ``Exact'' denotes the unmodified dense execution of the corresponding released ZPressor model. All values are reported in MiB.
    Reduction is computed as $1-\mathrm{\ours}/\mathrm{Exact}$ for the fusion-only peak from the underlying unrounded memory measurements; negative values therefore indicate additional routing and packing workspace at short contexts.
    }
    \label{tab:supp_memory_scaling}
    \setlength{\tabcolsep}{4pt}
    \small
    \begin{tabular}{l r rrr rr}
        \toprule
        &
        &
        \multicolumn{3}{c}{Fusion incremental peak}
        &
        \multicolumn{2}{c}{Full-model peak allocated} \\
        \cmidrule(lr){3-5}
        \cmidrule(lr){6-7}
        Backbone
        & Views
        & Exact
        & Ours
        & Reduction
        & Exact
        & Ours \\
        \midrule

        \multirow{5}{*}{MVSplat}
        & 36
        & 120.0
        & 162.8
        & $-35.6\%$
        & 3422.8
        & 3424.2 \\

        & 64
        & 163.5
        & 170.3
        & $-4.1\%$
        & 3442.8
        & 3444.1 \\

        & 128
        & 302.0
        & 182.3
        & $39.7\%$
        & 3490.9
        & 3492.0 \\

        & 192
        & 484.0
        & 164.0
        & $66.1\%$
        & 4194.5
        & 4195.9 \\

        & 256
        & 653.0
        & 170.8
        & $73.8\%$
        & 5570.5
        & 5572.1 \\

        \midrule

        \multirow{5}{*}{DepthSplat}
        & 36
        & 47.6
        & 60.0
        & $-26.0\%$
        & 2425.1
        & 2425.9 \\

        & 64
        & 47.6
        & 60.8
        & $-27.6\%$
        & 2459.1
        & 2458.5 \\

        & 128
        & 76.2
        & 64.4
        & $15.6\%$
        & 3413.5
        & 3413.8 \\

        & 192
        & 121.8
        & 58.3
        & $52.1\%$
        & 4789.4
        & 4789.6 \\

        & 256
        & 163.2
        & 60.4
        & $63.0\%$
        & 6165.5
        & 6165.8 \\

        \bottomrule
    \end{tabular}
\end{table*}
\begin{table*}[!t]
    \centering
    \caption{
    \textbf{Routing-load scaling for the MVSplat--ZPressor runtime benchmark.}
    Routing statistics are exact offline counts for the corresponding runtime-scaling workloads.
    $\widetilde N_c$ denotes the routed support-token length per cluster.
    Stride columns report the number of the six anchor clusters assigned
    to each stride.
    Cross-view latency is reported as Exact / \ours in ms.
    }
    \label{tab:supp_routing_load_scaling}
    \setlength{\tabcolsep}{4pt}
    \small
    \begin{tabular}{rrrrrrrr}
        \toprule
        Views
        & \shortstack{Mean support\\views/cluster}
        & \shortstack{Mean routed\\K/V length}
        & \multicolumn{3}{c}{Stride count}
        & QK Ret. $\downarrow$
        & \shortstack{Cross-view (ms)\\Exact / \ours} \\
        \cmidrule(lr){4-6}
        & & & $s_c=2$ & $s_c=4$ & $s_c=8$ & & \\
        \midrule
        36  & 5.00  & 2432.0 & 3 & 3 & 0 & 11.88\% & 17.86 / 4.64 \\
        64  & 9.67  & 2858.7 & 1 & 5 & 0 &  7.22\% & 31.98 / 4.90 \\
        128 & 20.33 & 3541.3 & 0 & 4 & 2 &  4.25\% & 62.95 / 5.33 \\
        192 & 31.00 & 2464.0 & 0 & 1 & 5 &  1.94\% & 93.95 / 4.60 \\
        256 & 41.67 & 2890.7 & 0 & 1 & 5 &  1.69\% & 125.28 / 4.80 \\
        \bottomrule
    \end{tabular}
\end{table*}
\begin{table*}[!t]
    \centering
    \caption{
    \textbf{Full routing decomposition at 256 context views.}
All variants use identical support views, K/V token counts, and QK
interactions, differing only in spatial phase assignment
($\kappa=5$ for layer cycling).
View Cov. measures the fraction of spatial phases covered across support
views within a cluster and layer, averaged equally over cluster--layer
pairs; Layer Cov. measures the corresponding coverage of each support
view across the six fusion layers.
Quality statistics are computed from per-scene $\Delta$PSNR relative to
Exact over $1{,}034$ scenes, with eight target views averaged within
each scene.
$N_{<-\tau}$ counts scenes with $\Delta$PSNR below $-\tau$ dB.
    }
    \label{tab:supp_routing_decomposition}
    \setlength{\tabcolsep}{4pt}
    \small
    \begin{tabular}{lrrrrrrrr}
        \toprule
        Routing
& \shortstack{View\\Cov.}
& \shortstack{Layer\\Cov.}
& \shortstack{Mean\\$\Delta$PSNR $\uparrow$}
& \shortstack{p5\\$\Delta$PSNR $\uparrow$}
& \shortstack{Worst\\$\Delta$PSNR $\uparrow$}
& $N_{<-.01}$ $\downarrow$
& $N_{<-.03}$ $\downarrow$
& $N_{<-.05}$ $\downarrow$\\
        \midrule
        Fixed & 2.41\% & 1.85\% & +0.0027 & -0.0204 & -0.1055 & 140 & 24 & 10 \\
        View-only & 74.11\% & 1.85\% & +0.0005 & -0.0064 & -0.0640 & 25 & 2 & 1 \\
        Layer-only & 2.41\% & 11.10\% & +0.0005 & -0.0154 & -0.1071 & 100 & 15 & 2 \\
        View+Layer & 74.11\% & 11.10\% & -0.0001 & -0.0063 & -0.0478 & 21 & 2 & 0 \\
        \bottomrule
    \end{tabular}
\end{table*}
\begin{table}[!t]
    \centering
    \caption{
    \textbf{Sensitivity to the layer-wise phase increment at 256 context views.}
Results are evaluated on the same fixed 200-scene subset used for the
K/V sensitivity study.
All settings use identical routed token counts and phase-coverage
statistics; only $\kappa$ is varied.
    }
    \label{tab:phase_increment}
    \setlength{\tabcolsep}{4pt}
    \begin{tabular}{crrrr}
        \toprule
        $\kappa$
        & \shortstack{Mean\\$\Delta$PSNR $\uparrow$}
        & \shortstack{p5\\$\Delta$PSNR $\uparrow$}
        & \shortstack{Worst\\$\Delta$PSNR $\uparrow$}
        & $N_{<-.01}$ $\downarrow$ \\
        \midrule
        1 & +0.0001 & -0.0045 & -0.0333 & 1 \\
        3 & -0.0002 & -0.0066 & -0.0239 & 2 \\
        5 & -0.0002 & -0.0059 & -0.0143 & 2 \\
        7 & -0.0002 & -0.0055 & -0.0090 & 0 \\
        \bottomrule
    \end{tabular}
\end{table}

\subsection{Extended Scaling Results}

\subsubsection{Runtime and Memory Scaling}

We provide the complete runtime and memory measurements underlying the
scaling analysis (\cref{fig:runtime_scaling,fig:memory_scaling,tab:scaling_summary}) in the main paper.
\cref{tab:supp_runtime_scaling} reports cross-view fusion, ZPressor~\cite{zpressor}, encoder, and end-to-end latency for both MVSplat~\cite{mvsplat} and DepthSplat~\cite{depthsplat} over increasing context lengths.
\cref{tab:supp_memory_scaling} reports the corresponding fusion-only incremental peak memory together with full-model peak allocated memory.

Across both backbones, dense cross-view fusion becomes progressively more expensive as the context grows, whereas the routed latency changes only modestly.
The resulting end-to-end gain is larger on MVSplat~\cite{mvsplat} because cross-view fusion constitutes a larger fraction of its overall runtime.
DepthSplat~\cite{depthsplat} spends a larger fraction of computation in unchanged components, so similar fusion-level scaling produces a smaller system-level improvement.

The memory results in \cref{tab:supp_memory_scaling} show the same distinction between the modified fusion path and the complete model.
Our routing and packing introduce additional overhead at short contexts, but the fusion-only incremental peak remains approximately stable as more views are added.
Dense fusion memory instead grows with context length, so the relative memory advantage of \ours becomes increasingly significant.
The full-model allocated peak changes little between Exact and \ours, indicating that the overall peak is dominated by components outside the cross-view fusion stage.

\subsubsection{Routing-Load Scaling}

To examine the mechanism behind the nearly flat routed latency, \cref{tab:supp_routing_load_scaling} reports exact routing statistics for the MVSplat~\cite{mvsplat} runtime scaling benchmark.

From 36 to 256 context views, the dense support-token count grows by $8.33\times$, whereas the mean routed K/V length remains between $2432.0$ and $3541.3$ tokens.
This behavior follows directly from the view-normalized routing rule: as more support views are assigned to each anchor, the selected spatial stride becomes coarser, reducing the number of retained locations per view while keeping every support camera represented. Accordingly, QK retention decreases from $11.88\%$ to $1.69\%$, while the routed cross-view latency remains within $4.60$--$5.33$ ms.

The routed workload is not strictly monotonic because stride selection is discrete and support views are not distributed perfectly evenly across anchor clusters.
At 128 views, four of the six clusters still use $s_c=4$ and only two use $s_c=8$, producing the largest mean routed length in the sweep.
By 192 views, five clusters use $s_c=8$, reducing the routed workload despite the larger number of support views.
The stride distribution therefore explains the modest non-monotonicity in both routed token count and latency while preserving the overall bounded scaling behavior.
The $1.69\%$ QK retention reported here corresponds to the runtime-scaling workload at 256 views. The corresponding evaluation-set global ratio over all $1{,}034$ scenes is $1.85\%$.

\subsection{Routing Ablations and Sensitivity}

\subsubsection{Full Routing Decomposition}

The main paper reports a compact comparison of Fixed, View-only, Layer-only, and View+Layer routing in \cref{tab:routing_decomposition}.
Here, \cref{tab:supp_routing_decomposition} expands this comparison with phase coverage statistics and additional lower-tail quality measurements. All variants use identical support views, K/V token counts, and QK interactions, isolating the effect of sampling offsets.

The coverage statistics confirm that the ablations isolate the intended
routing dimensions: View-only increases phase coverage across support
views while leaving layer-wise coverage unchanged, whereas Layer-only
changes the coverage across fusion layers without staggering support views.
View-wise staggering provides the dominant robustness improvement.
Compared with Fixed routing, View-only substantially improves both p5
and worst-case $\Delta$PSNR and reduces the number of scenes below the
reported degradation thresholds.

Layer-only routing provides a weaker improvement.
Although it reduces several failure counts, its worst case is slightly worse than Fixed, showing that varying spatial locations only across fusion layers cannot replace distributing the available evidence across support views.
Adding layer cycling to View-only leaves the p5 behavior nearly unchanged but further improves the extreme tail. In particular, the worst-case degradation improves from $-0.0640$ dB to $-0.0478$ dB and $N_{<-.05}$ decreases from $1$ to $0$.
These results support the default View+Layer design: view-wise staggering is the primary source of robustness, while layer-wise variation provides a complementary refinement for difficult scenes.

\subsubsection{Sampling Offset Sensitivity}

Having established that varying the sampling offsets across fusion layers improves the extreme tail, we examine whether this benefit depends on the particular layer-wise increment $\kappa$.
\cref{tab:phase_increment} compares several odd increments while keeping the routed token count, support views, and all other routing settings fixed.

For the power-of-two strides used in our experiments, $s_c^2$ is also a
power of two.
Any odd $\kappa$ is therefore coprime to $s_c^2$, so repeatedly adding $\kappa$ modulo $s_c^2$ gives a maximal-length phase cycle.
Changing $\kappa$ consequently changes the order in which sampling offsets are visited, but not the number of distinct phases that can be reached. This is consistent with the identical phase-coverage statistics observed for all tested settings.

The reconstruction results show that the particular traversal order has only a minor effect.
Across the tested increments, the mean $\Delta$PSNR remains essentially unchanged, and the p5 values stay within a narrow range.
The worst case varies somewhat more, but remains small for all settings and does not correspond to a consistent advantage in the average or p5 metrics.
In particular, $\kappa=5$ is not selected as the best-performing value in this sweep.
We use it as a fixed offset because it provides the intended layer-wise phase variation while behaving similarly to the other odd choices.
The effectiveness of the routing therefore comes from varying spatial evidence across fusion layers, rather than from tuning a specific phase increment.
%%%%%%%%% 

\end{document}